\documentclass[letterpaper, preprint, paper,11pt]{AAS}	% for preprint proceedings

\usepackage{bm}
\usepackage{amssymb}
\usepackage{amsmath}

\usepackage{amsfonts}

\usepackage[colorlinks=true, pdfstartview=FitV, linkcolor=black, citecolor= black, urlcolor= black]{hyperref}
\usepackage{overcite}
\usepackage{booktabs}
\usepackage{multirow}
\usepackage{footnpag}			      	% make footnote symbols restart on each page
\usepackage{soul}
\usepackage{array}
\usepackage{subcaption}

\usepackage[normalem]{ulem}

\PaperNumber{25-891}

\begin{document}

%\title{BIO-INSPIRED {CONTROL} STRATEGY FOR SPACE MANIPULATOR SYSTEMS BASED ON ANIMAL RIGHTING REFLEXES}

\title{Adapting Biological Reflexes for Dynamic Reorientation in Space Manipulator Systems}

\author{Daegyun Choi\thanks{Postdoctoral Researcher, Department of Aerospace Engineering \& Engineering Mechanics, University of Cincinnati, Cincinnati, OH 45221, USA.}, 
Alhim Vera\thanks{Graduate Student, Department of Aerospace Engineering \& Engineering Mechanics, %veragoaa@mail.uc.edu
University of Cincinnati, Cincinnati, OH 45221, USA.},  and
Donghoon Kim\thanks{Assistant Professor, Department of Aerospace Engineering \& Engineering Mechanics, University of Cincinnati, Cincinnati, OH 45221, USA.}
% Daegyun Choi\thanks{Postdoctoral Researcher, Department of Aerospace Engineering \& Engineering Mechanics, choidg@ucmail.uc.edu},
% \ and Donghoon Kim\thanks{Assistant Professor, Department of Aerospace Engineering \& Engineering Mechanics, kim3dn@ucmail.uc.edu.}
}

\maketitle{}

\begin{abstract}

Robotic arms mounted on spacecraft, known as space manipulator systems (SMSs), are critical for enabling on-orbit assembly, satellite servicing, and debris removal. However, controlling these systems in microgravity remains a significant challenge due to the dynamic coupling between the manipulator and the spacecraft base. This study explores the potential of using biological inspiration to address this issue, focusing on animals, particularly lizards, that exhibit mid-air righting reflexes. Based on similarities between SMSs and these animals in terms of behavior, morphology, and environment, their air-righting motion trajectories are extracted from high-speed video recordings using computer vision techniques. These trajectories are analyzed %using 
within a multi-objective optimization framework to identify the key behavioral goals and assess their relative importance. The resulting motion profiles are then applied as reference trajectories for SMS control, with baseline controllers used to track them. The findings provide a step toward translating evolved animal behaviors into interpretable, adaptive control strategies for space robotics, with implications for improving maneuverability and robustness in future missions.
\end{abstract}

\section{Introduction}
\subsection{Background}

Space manipulator systems (SMSs), composed of spacecraft equipped with robotic arms, are essential for enabling in-space servicing, assembly, and manufacturing, as well as debris removal. Remarkable examples of SMSs that have demonstrated their capabilities in space include the Engineering test satellite no. 7 (ETS-VII), developed by the National Space Development Agency of Japan \cite{yoshida2003ets} and autonomous space transport robotic operations (ASTRO) system, developed by Boeing Integrated Defense Systems \cite{shoemaker2003orbital}.
The successful demonstrations of both SMSs have shown the viability of autonomous robotic operations in orbit. These missions highlight the increasing need for high-precision, energy-efficient, and adaptable manipulation strategies to support the growing complexity of future space missions.

In microgravity, SMSs face unique challenges despite considerable progress in operating and testing them on orbit. Unlike terrestrial manipulators anchored to rigid bases, robotic arms in space operate from a floating spacecraft. This introduces strong dynamic coupling—any motion of the arm(s) induces a reactive motion on the base spacecraft, affecting overall system stability and control accuracy \cite{rodrigues2021}. Additionally, constraints on fuel, computational power, and safety require appropriate motion planning strategies that can execute with high efficiency and reliability.

Traditional optimal control techniques have focused on minimizing relative distances to targets \cite{qiao2019motion} or reducing energy consumption \cite{app11052346}, while considering several factors, such as collision avoidance, singularity avoidance, and the robotic arm's limited capabilities. However, these methods are computationally expensive and can be difficult to implement in real-time operations. Furthermore, they may lack robustness under uncertainties in real-world space environments. 
On the other hand, some researchers have studied reinforcement learning and other artificial intelligence-based techniques for SMS control to overcome the traditional methods' drawbacks \cite{Blaise2023}. Such approaches offer a more adaptive solution space and greater robustness against uncertainties \cite{rlli2022}. However, their black-box nature and lack of interpretability, making it difficult for control engineers to understand their decision-making, present challenges for implementation in high-assurance and safety-critical space robotic systems \cite{LONGO2024102301}.

\subsection{Problem Statement}

\begin{figure}[tbp]
    \centering
    \includegraphics[width=0.9\linewidth]{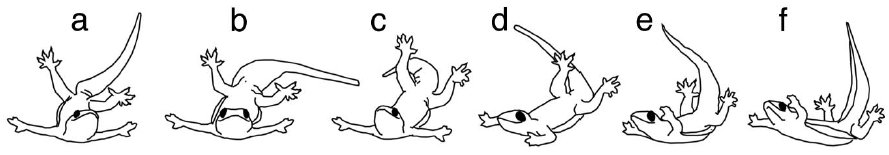}
    \caption{Sequence of Aerial Righting of Gecko. Adapted from \cite{jusufi2008active} Copyright (2008) National Academy of Sciences.}
    \label{fig:lizard_free_fall}
\end{figure}

To overcome these limitations, this research investigates a biologically inspired operational strategy that leverages optimization principles observed in nature. Animals, such as cats, squirrels, and especially geckos, have evolved mid-air righting reflexes that achieve stability, efficiency, and safety under dynamic conditions. For example, cats dynamically twist their bodies %for righting and safe landing 
to right themselves and land safely during unexpected falls \cite{kane1969dynamical}, and squirrels utilize their tails to stabilize their posture during free-fall situations \cite{SICB2021_squirrel}. In addition, as shown in Figure~\ref{fig:lizard_free_fall}, lizards reorient in mid-air by swinging their tails to redistribute angular momentum \cite{jusufi2008active,josufi2011aerial}, a mechanism that parallels momentum exchange in SMSs. These strategies result from natural optimization processes that balance various objectives, such as stability, agility, and energy conservation.

This work hypothesizes that evolved animal behaviors can inform the design of interpretable and robust control strategies for SMSs in microgravity. To validate this hypothesis, the research focuses on biomechanical modeling of lizard motion and the extraction of tail-body coordination patterns using computer vision techniques. The extracted trajectories are then used to identify multiple objectives and their corresponding weighting factors, providing a mathematical interpretation of the animals' evolved behaviors. These motion trajectories serve as reference inputs to a dynamic model of an SMS, enabling evaluation of its behavior and comparison with animal motion. These insights lay the groundwork for developing adaptive and efficient control strategies for future autonomous SMS operations.

\section{Proposed Methodology and Results}
The proposed methodology aims not only to understand the animals' evolved behaviors by identifying similarities with SMSs and extracting motion trajectories, but also to implement these trajectories into the SMS dynamics model as reference inputs. These reference trajectories guide the controller in generating appropriate control inputs while allowing for the analysis of the underlying behavioral objectives.

The overall process consists of three main steps:

\subsubsection{\textbf{{Step 1: Identification of Similarities and Objective Formulation}}}

The first step involves identifying behavioral, morphological, and environmental similarities between biological systems and SMSs. Motivated by the widely supported view that evolved animal behaviors embody natural optimization principles, an objective function containing multiple objectives and weighting factors is formulated. This study focuses on a single robotic arm-mounted spacecraft configuration, modeled after real-world missions, such as ETS-VII and ASTRO. Building on these comparisons, the righting reflex observed in lizards is chosen for deeper investigation.

\subsubsection{\textbf{Step 2: Motion Trajectory Extraction via Computer Vision}}

To model animals' righting reflexes, motion trajectories are extracted from recorded videos of lizards in free-fall using various computer vision techniques. Key body segments involved in mid-air reorientation are identified through this analysis, enabling the construction of a biomechanical model of the observed behavior.

\subsubsection{\textbf{{Step 3: Integration and Control Evaluation in SMS (In progress)}}} 

The extracted trajectories and identified body segment dynamics are then used to infer the objectives and weighting factors within the previously defined optimization framework, focusing on metrics, such as safety, stabilization, and efficiency. These trajectories are implemented as reference paths in the SMS dynamics model. Controllers, such as proportional-integral-derivative, linear-quadratic regulator, or model predictive control, are planned for this task. By observing the SMS's behavior under these control schemes, the influence of different weighting factors in the objective function can be evaluated and analyzed.

\subsection{Identification of Similarities and Objective Formulation}

\subsubsection{\textbf{Behavioral Similarities}}
In nature, animal motion reflects evolved strategies for surviving in dynamic and uncertain environments, often balancing competing objectives, such as stability, agility, and energy efficiency \cite{jusufi2011lizard, chang2011active}. 
During free-fall scenarios (Figure~\ref{fig:lizard_free_fall}), lizards exhibit active body adjustments that reflect these priorities. Such behaviors can be interpreted as solutions to a multi-objective optimization problem influenced by environmental constraints and internal priorities \cite{animal_behavior}: 
\begin{equation}
    \min J = w_1 \phi_\text{safety} + w_2 \phi_\text{stability} + w_3 \phi_\text{efficiency}.
    \label{eq:cost} 
\end{equation} 
Here, $w_i$ represent the weighting coefficients for each behavioral component, $\phi_\text{safety}$ reflects critical survival actions (e.g., protecting the head), $\phi_\text{stability}$ reflects postural control or reorientation, and $\phi_\text{efficiency}$ penalizes unnecessary energy expenditure. These weights shift dynamically depending on task demands or environmental stimuli. 

Similarly, SMS operations can be formulated as a multi-objective optimization problem, where performance metrics like collision avoidance, energy usage, and disturbance minimization must be balanced under dynamic coupling and physical constraints. Like animals, SMSs must adapt to disturbances while maintaining task success. The behavioral analogies between animals and SMSs are summarized in Table~\ref{tab:righting_analogy}.

\begin{table}[tbp]
\centering
\caption{Behavioral Analogy}
\begin{tabular}{p{2cm}|p{3.5cm}|p{4cm}|p{3.5cm}}
\hline
\textbf{Aspect} & \textbf{Lizards} & \textbf{SMSs} & \textbf{Analogy} \\
\hline
\textbf{Safety} & Instinctively reorient during a fall using tail and limbs to reduce impact risk & Uses onboard attitude control (e.g., thrusters, reaction wheels) to correct orientation during unexpected motion & Both respond rapidly to restore a safe orientation when destabilized \\
%\hline
\textbf{Stabilization} & Adjust body posture mid-air through coordinated limb and tail movement & Coordinates base and arm movements to maintain or regain system stability & Both rely on internal coordination to stabilize themselves during motion \\
%\hline
\textbf{Efficiency} & Executes reorientation with minimal, well-timed effort & Employs optimized control laws to correct attitude with minimal energy use & Both achieve orientation correction without excessive energy expenditure \\
\hline
\end{tabular}
\label{tab:righting_analogy}
\end{table}

While certain behavioral analogies exist between lizards and SMSs, notable differences remain, particularly in environmental context and movement timescales. SMSs typically perform slow, deliberate motions over extended durations during proximity operations, whereas lizards execute righting reflexes in free-fall conditions that last less than one second and involve significant vertical displacement \cite{jusufi2008active}. Although aerodynamic drag may play a role in lizard reorientation, it is considered negligible here due to the short duration of the fall, the small effective surface area, and the lack of fur, unlike %in 
animals,such as cats or squirrels.

As shown in Figure~\ref{fig:lizard_free_fall}, lizards use rapid tail motions to reorient themselves mid-air, movements far quicker than those achievable by current SMS actuation systems. To bridge this gap, the planned work involves scaling and modifying the extracted lizard motion trajectories to fit within the physical and dynamic constraints of SMS. This approach enables a practical transfer of biologically inspired strategies to robotic systems operating in space.

% These parallels reinforce the core hypothesis of this study: biologically optimized motion patterns can inform the development of interpretable, efficient, and robust control strategies for space robotics.

\subsubsection{\textbf{Morphological Similarity}}
This work mainly concentrates on an SMS configuration featuring a single robotic arm, such as those found in ETS-VII and ASTRO. Among various animals known for their righting reflexes, lizards present a particularly strong geometric analogy to actual SMS platforms. To examine this morphological similarity, one compares the segment proportions between lizards and SMSs. As shown in Table~\ref{tab:lizard_robot_proportion}, the relative segment lengths exhibit notable alignment. For instance, a lizard’s tail accounts for approximately 57\% of its total body length, which is comparable to the reach of a robotic arm relative to one side of the spacecraft base. This geometric similarity supports the suitability of lizards as a biological analog for analyzing SMS behavior. 

\begin{figure}[ht]
    \centering
    \begin{subfigure}[b]{0.4\textwidth}
        \includegraphics[width=\linewidth]{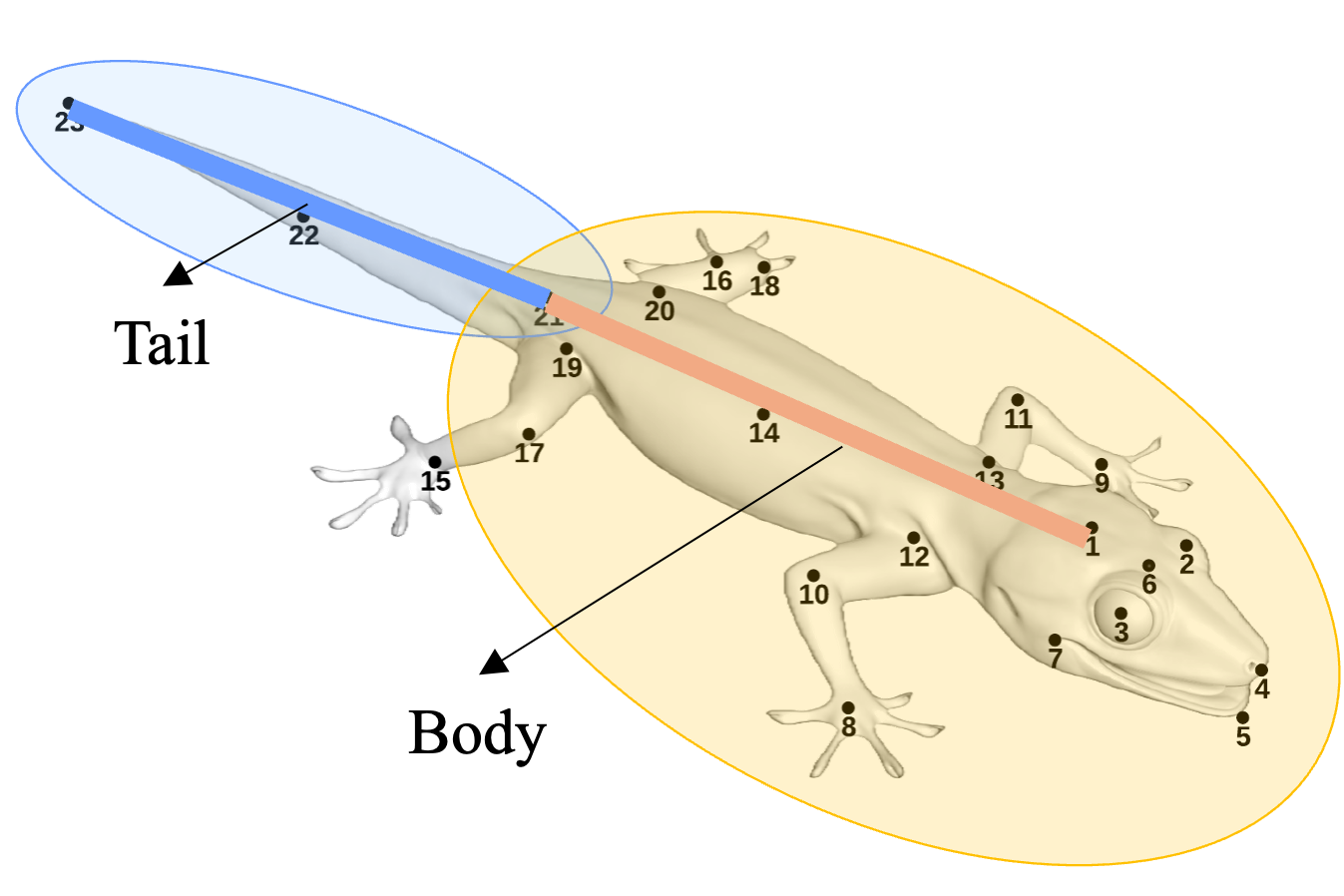}
        \caption{Lizard}
        \label{fig:lizard}
    \end{subfigure}
    \begin{subfigure}[b]{0.29\textwidth}
        \includegraphics[width=\linewidth]{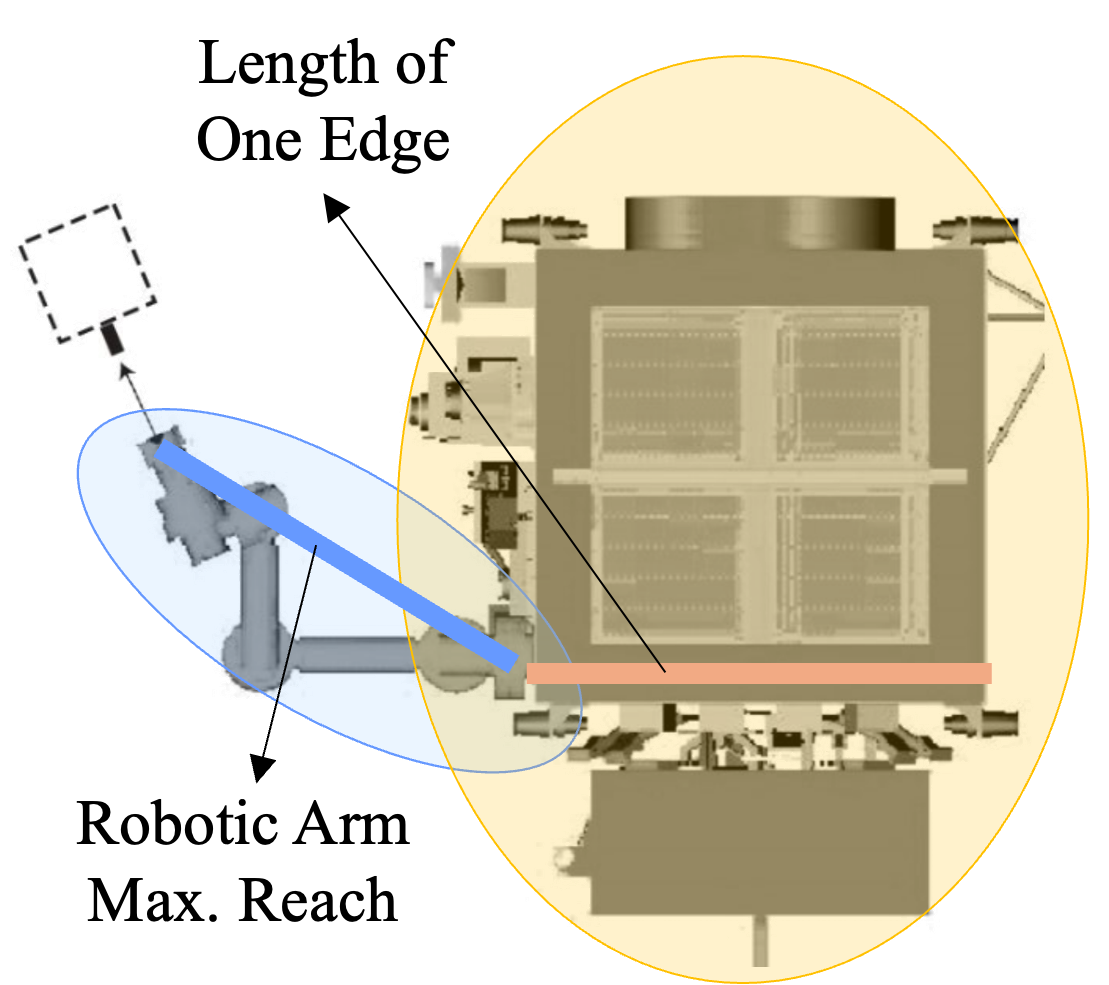}
        \caption{ETS-VII \cite{yoshida2003ets}}
        \label{fig:ETS-VII}
    \end{subfigure}
    \begin{subfigure}[b]{0.29\textwidth}
        \includegraphics[width=\linewidth]{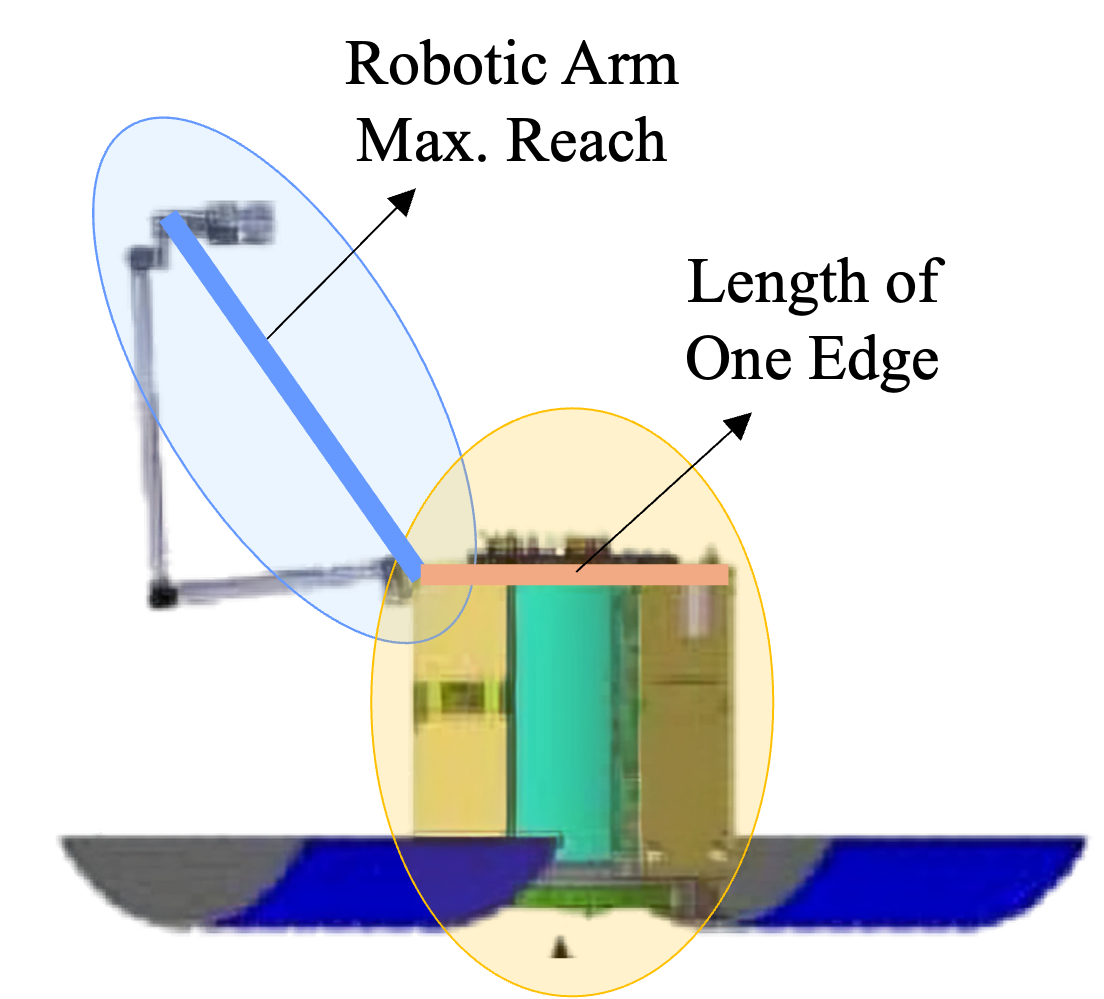}
        \caption{ASTRO \cite{shoemaker2003orbital}}
        \label{fig:ASTRO}
    \end{subfigure}
    \caption{Visual Comparison of Lizard and SMSs}
    \label{fig:body_tail_relation}
\end{figure}

Figure~\ref{fig:body_tail_relation} further illustrates this relationship: the body (in orange) and tail (in blue) segmentation in lizards is directly compared with two satellite-manipulator systems. In both biological and engineered systems, the elongated appendage (tail or robotic arm) can serve as an actuator for body reorientation and interaction with the environment.

However, in terms of mass, although the mass ratio of the robotic arm in SMSs differs from that of lizards, this ratio does not account for the payload mass in SMSs. This available margin for additional mass on the robotic arm allows us to explore various payloads during SMS operation in the analysis.

\begin{table}[ht]
    \caption{Morphological Analogy}
    \centering
    \begin{tabular}{p{4cm}|p{1.8cm}|p{2.3cm}|p{4.9cm}}
        \hline
        \textbf{Aspect} & \textbf{Lizards} \cite{Jayne1999} & \textbf{SMSs} \cite{yoshida2003ets,shoemaker2003orbital} & \textbf{Analogy} \\ \hline
        Primary body segments & 2 (body \& tail) & 2 (spacecraft \& robotic arm) & {Shown in} Figure~\ref{fig:body_tail_relation} \\ %\hline
        Length ratio (tail or %robotic 
        arm / total length) & 0.31 -- 0.65 & 0.35 -- 0.43 & {Comparable segment proportions} \\
       % \hline
        Mass ratio (tail or %robotic 
        arm / {body or base }%total 
        mass)& $\sim$ 0.2 & {$\geq$ 0.05} & {Can be comparable when accounting for SMS payload mass} \\
        %\hline
        Degree of freedom {(DOF)} & $\sim\infty$ & 12 (6 + 6) & {Depends on the motion of interest in lizards} \\ %\hline
        Detailed body segments & $\sim\infty$ & 7 (1 + 6) & {Approximation depends on modeling objectives} \\ \hline        
    \end{tabular}
    \label{tab:lizard_robot_proportion}
\end{table}

% \begin{table}
%     \centering
%     \caption{Environmental Analogy}
%     \label{tab:env_analogy}
%     \begin{tabular}{p{2.3cm}|p{4.3cm}|p{3.7cm}|p{2.9cm}}
%     \hline
%         \textbf{Aspect} & \textbf{Lizards} & \textbf{SMSs} & \textbf{Analogy}  \\ \hline
%         Air drag effect & Present, but minimal due to small size and rapid falling & Dependent on orbit, but typically negligible & Both experience negligible air drag \\ \hline
%         Gravity effect* & Microgravity during free fall conditions & Microgravity & Both experience microgravity \\ \hline
%         \multicolumn{4}{l}{* Some studies\cite{barabanov2015gecko, morozova2019reptiles} show that lizards' self-righting behavior in space is similar to that on Earth.}\\
%     \end{tabular}
% \end{table}

From a modeling perspective, both lizards and SMSs can be treated as multi-body systems with jointed segments. Their dynamics are governed by the standard equations of motion \cite{Antonello2019}: 
\begin{equation}
    M(\mathbf{q})\ddot{\mathbf{q}} + C(\mathbf{q}, \dot{\mathbf{q}})\dot{\mathbf{q}} + {\mathbf{g}}({\bf q})= \bm{\tau} + \bm{\tau}_d, 
    \label{eq:model} 
\end{equation} 
where $M \in \mathbb{R}^{m \times m}$ is the mass/inertia matrix, $C \in \mathbb{R}^{m \times m}$ is the Coriolis and centrifugal matrix,  ${\mathbf{g}}\in\mathbb{R}^{m}$ is a vector related to the gravity term, $\mathbf{q} \in \mathbb{R}^{m}$ represents the joint configuration, $\bm{\tau} \in \mathbb{R}^{m}$ denotes the applied joint torques, and $\bm{\tau}_d \in \mathbb{R}^{m}$ indicates the disturbances, like air drag, acting on the system. Note that the gravity term is generally assumed to be zero for SMSs. When external force and torque acting on the system are zero, the system exhibits momentum conservation, a key principle underlying both lizard mid-air motion and SMS motion. This property implies that movement in one segment, such as a lizard’s tail, induces reactive motion in another, like the torso, mirroring how the motion of an SMS’s robotic arm perturbs the spacecraft base. 
In fact, the number of DOF and detailed body segments differs because lizards have a flexible body. However, these can be appropriately modeled depending on the specific motion of interest, which will be discussed in Step 2.

\subsubsection{\textbf{Environmental Similarity}}
Even though lizards experience less aerodynamic drag in mid-air, gravity can still introduce behavioral discrepancies. This work mainly considers free-fall conditions for observing lizards' behavior, as these conditions closely resemble the microgravity environments in which SMSs operate. In fact, empirical studies \cite{barabanov2015gecko, morozova2019reptiles} support using free-fall settings to examine lizards' righting reflex behavior.

These studies show that geckos exhibit skydiving-like limb extensions and tail actuation during free-fall in space, closely mimicking their terrestrial mid-air responses. This reinforces the hypothesis that mid-air animal behaviors, optimized for gravity-dominated environments, remain valid in microgravity or near-zero-g contexts. Despite differences in mass, air drag, and actuation dynamics, this environmental similarity, along with behavioral and morphological parallels, suggests that biologically inspired movement primitives can inform SMS operation strategies.

\begin{table}[t]
    \centering
    \caption{Environmental Analogy}
    \label{tab:env_analogy}
    \begin{tabular}{p{2.3cm}|p{4.3cm}|p{3.7cm}|p{2.9cm}}
    \hline
        \textbf{Aspect} & \textbf{Lizards} & \textbf{SMSs} & \textbf{Analogy}  \\ \hline
        Air drag effect & Present, but minimal due to small size and rapid falling & Dependent on orbit, but typically negligible & Both experience negligible air drag \\ %\hline
        Gravity effect* & Microgravity during free fall conditions & Microgravity & Both experience microgravity \\ \hline
        \multicolumn{4}{l}{\footnotesize * Some studies \cite{barabanov2015gecko, morozova2019reptiles} show that lizards' self-righting behavior in space is similar to that on Earth.}\\
    \end{tabular}
\end{table}

These observations form the theoretical basis for this work: translating righting reflexes observed in lizards into operational strategies for SMSs. By extracting motion patterns from animal data and understanding their governing dynamics, the goal is to build a foundation for interpretable and adaptive SMS control frameworks based on biological intelligence.

% These observations form the theoretical foundation for this work: translating the righting reflexes observed in lizards into operational strategies for SMSs. By extracting motion patterns from animal data and understanding their underlying dynamics, we aim to establish a foundation for interpretable and adaptive SMS control frameworks based on biological intelligence.

\subsection{Motion Trajectory Extraction via Computer Vision}\label{sec:MT}
% To address the limitations of traditional control strategies and AI-based methods for SMSs, this work proposes a bio-inspired methodology that draws from the righting reflex behaviors observed in animals—specifically lizards. These reflexes are viewed as evolved solutions to stabilize body orientation in an efficient way under dynamic conditions, providing insights that can be applied to SMSs operating in microgravity environment. 

To efficiently obtain, evaluate, and analyze the motion data, this study constructs a computer vision-based pipeline focused on extracting the lizard's motion trajectories. The structured pipeline for motion pattern extraction, developed in the authors' previous work \cite{Vera2025}, consists of five steps:
i) Segmentation is performed using the segment anything model \cite{kirillov2023segment} to focus on the object of interest in the video frames; 
ii) Pose estimation is conducted with ViTPose++ \cite{xu2022vitpose} to estimate the pose information of the object based on keypoints around it; 
iii) Tracking is handled by CoTracker \cite{karaev2023cotracker} to continuously track the detected keypoints;
iv) Temporal smoothing and identity re-association are implemented using a centroid-based optimization strategy, ensuring reliable multi-frame trajectories for analysis;
v) Transformation from 2-dimensional (2D) image coordinates to 3D world coordinates is conducted to reconstruct the 3D information and compute the motion trajectories.

\subsubsection{\textbf{Body Segment Identification}}

\begin{figure}[t]
    \centering
    \includegraphics[width=0.8\linewidth]{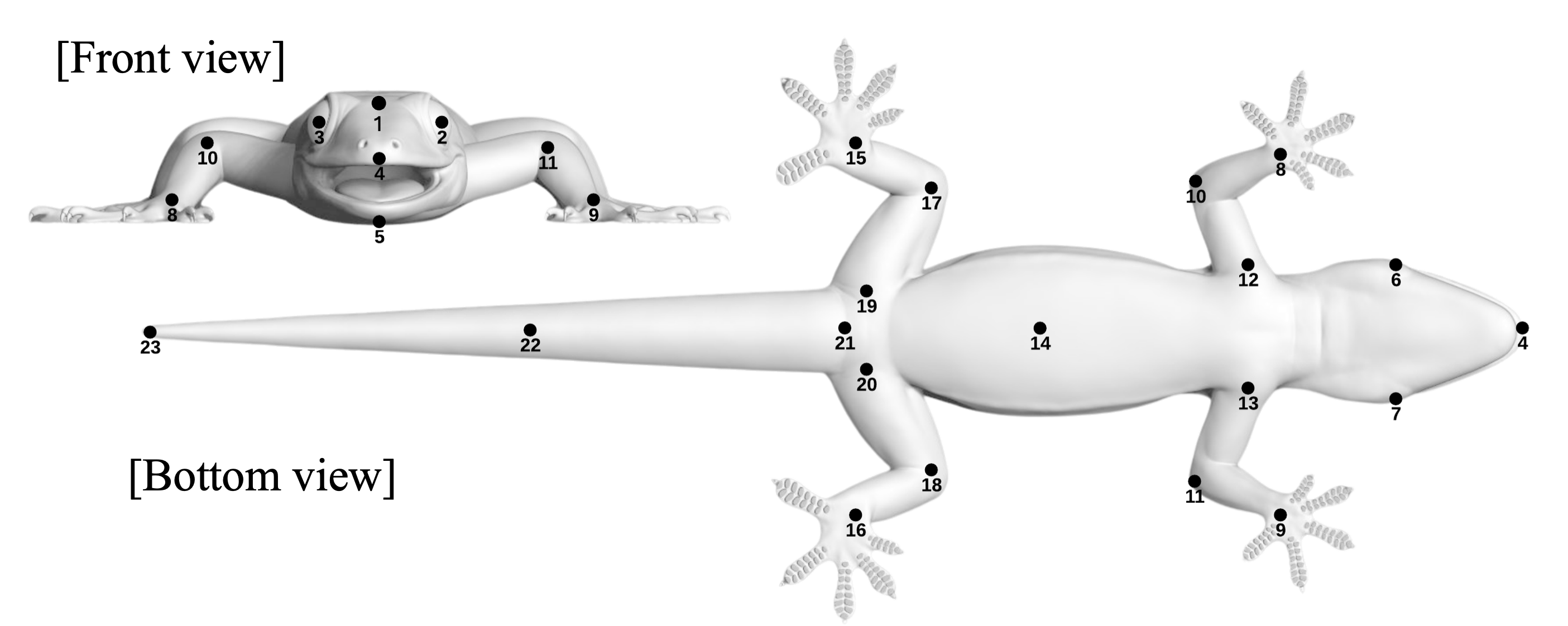}
    \caption{Annotated 23 Keypoints in Lizard and 14 Anatomical Body Segments}
    \label{fig:lizard_keypoints_ref}
\end{figure}

To obtain effective motion trajectories that describe righting reflexes, this work begins by identifying the segments and keypoints of lizard anatomy. 
The anatomical structure is extracted from the Animal Kingdom dataset. As shown in Figure~\ref{fig:lizard_keypoints_ref}, each of the 23 keypoints corresponds to a distinct anatomical location critical for dynamic modeling. With these keypoints, motions across approximately 14 body segments can be observed.
\begin{figure}
    \centering
    \includegraphics[width=0.9\linewidth]{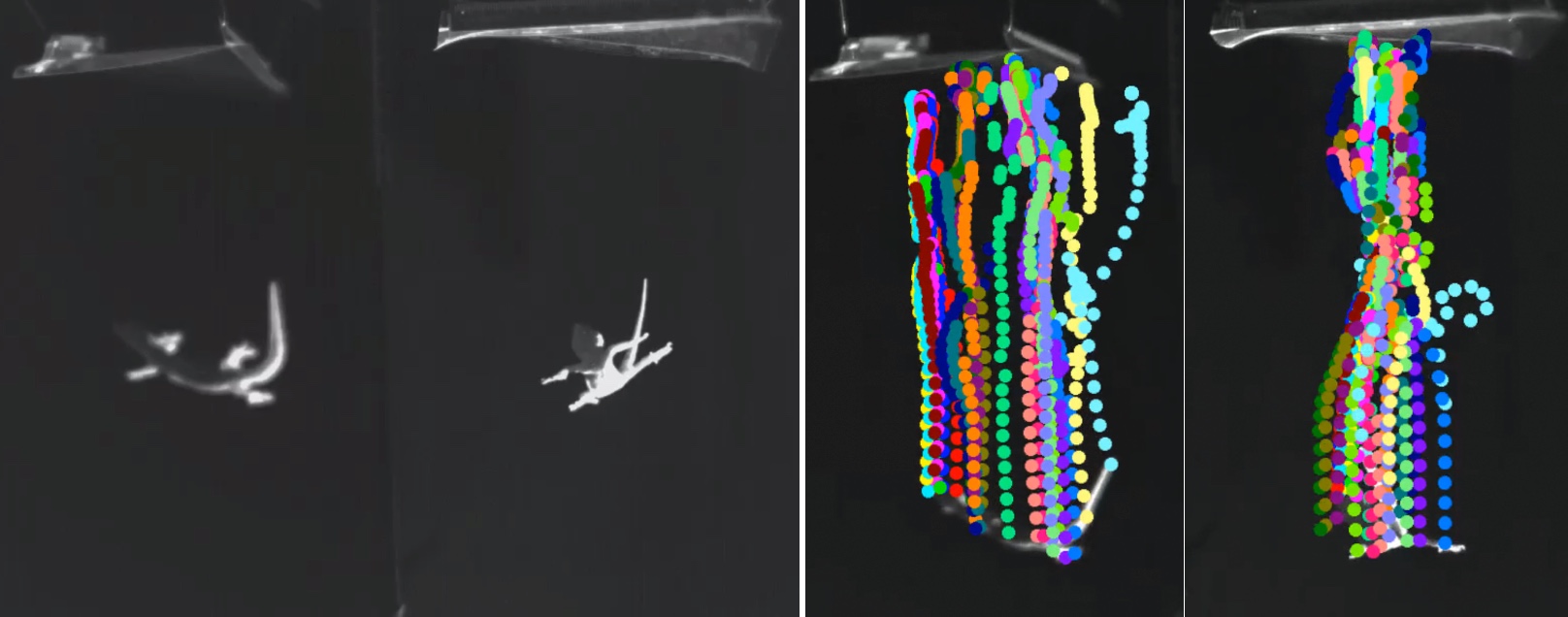}
    \caption{Capture of the Original Video \cite{Jusufi2010} (left) and the Keypoints Extraction (right)}
    \label{fig:keypoints_extraction}
\end{figure}

To evaluate the feasibility of %the 
motion trajectory extraction, each individual keypoint is identified from real free-fall {high-speed} video recordings of a lizard % at 30 frames per second 
\cite{Jusufi2010}, as shown in Figure~\ref{fig:keypoints_extraction}. Each extracted keypoint is then analyzed in terms of its movement and visibility, as presented in Table~\ref{tab:keypoint_motion}. Due to resource constraints, only a limited number of videos are processed; however, additional data will be collected to improve statistical confidence. The movement analysis quantifies the frame-to-frame displacement of each keypoint using Euclidean distance in pixel space.
%:
% \begin{equation}
%     M_{i,t} = \sqrt{(x_{i,t} - x_{i,t-1})^2 + (y_{i,t} - y_{i,t-1})^2}
%     \label{eq:euclidean_distance}
% \end{equation}
% where $(x_{i,t}, y_{i,t})$ represents the 2D coordinates of point $i$ at frame $t$. 

The average movement for each point is %then 
computed across all frames, and visibility is calculated as the percentage of frames in which the point is successfully tracked. 
% \begin{equation}
%     V_{i} = \frac{1}{T}\sum_{t=1}^{T} v_{i,t} \times 100\%
%     \label{eq:average_movement}
% \end{equation}
% where $v_{i,t}$ is a binary indicator (1 if visible, 0 if occluded) and $T$ is the total number of frames.
% The analysis pipeline processes video data at 30 fps through a comprehensive motion extraction framework. Each keypoint's position is tracked frame-by-frame using computer vision techniques, with positions stored as (x,y) coordinates in pixel space. For each frame $t$, the Euclidean distance between consecutive positions is computed, providing a measure of instantaneous movement. The instantaneous velocity is derived as $V_{i,t} = M_{i,t}/\Delta t$ where $\Delta t = 1/30$ seconds, enabling analysis of movement dynamics. 
To facilitate comparative analysis across different anatomical regions, movements are normalized relative to the maximum observed movement across all points. 
% The analysis incorporates anatomical segmentation, grouping points into functional units (head, body, tail, legs) to evaluate segment-level movement patterns and their contribution to the overall motion.

Notably, points along the tail (21–23), hips (19–20), and knees (17–18) show high movement and visibility, highlighting their role in dynamic body reorientation. In contrast, points near the mouth (4-7) and eye (2-3) show low movement and frequent occlusion, making them less useful for reliable modeling.
\begin{table}[tb]
    \centering
    \caption{Average Movement and Visibility of Keypoints}
    % \renewcommand{\arraystretch}{1.2}
    % \scriptsize
    \begin{tabular}{cccc}
        \hline
        \textbf{Keypoints} & {\textbf{Point name}} & {\textbf{Average movement (pixel)}} & {\textbf{Visibility (\%)}} \\
        \hline
        1 & Neck & 0.53 &  22.10  \\
        2 & Eye\_Left & 0.24 &  13.20   \\
        3 & Eye\_Right & 0.44 &  18.40   \\
        4 & Mouth\_Front\_Top & 0.24 &16.20 \\
        5 & Mouth\_Front\_Bottom & 0.72 &14.70 \\
        6 & Mouth\_Back\_Right & 0.65 &  19.90 \\
        7 & Mouth\_Back\_Left & 0.35 &  19.10 \\
        8 & Wrist\_Right & 0.53 &  22.10 \\
        9 & Wrist\_Left & 1.86 &  25.00 \\
        10 & Elbow\_Right & 1.22 &  24.30 \\
        11 & Elbow\_Left & 0.51 & 22.10 \\
        12 & Shoulder\_Right & 1.33 &  22.80 \\
        13 & Shoulder\_Left & 0.50 &  22.10 \\
        14 & Torso\_Mid\_Back & 0.79 &  14.00\\
        15 & Ankle\_Right & 1.41 &  23.50\\
        16 & Ankle\_Left & 1.80 & 33.80 \\
        17 & Knee\_Right & 1.38 &  30.10 \\
        18 & Knee\_Left & 2.12 &  27.90 \\
        19 & Hip\_Right & 0.95 &  27.90 \\
        20 & Hip\_Left & 4.35 & 30.10 \\
        21 & Tail\_Top\_Back & 0.66 &  22.10 \\
        22 & Tail\_Mid\_Back & 0.96 & 13.20 \\
        23 & Tail\_End\_Back & 2.67 &  18.40 \\
        \hline
    \end{tabular}
    \label{tab:keypoint_motion}
\end{table}

To evaluate the temporal reliability of keypoints during dynamic motion, a stability analysis based on frame-wise time series data is performed. This analysis helps determine which anatomical landmarks remain consistently trackable, which %ones 
drift erratically, and which are frequently lost due to occlusion. These insights are essential for obtaining continuous trajectory information.

Table~\ref{tab:keypoint_stability} summarizes several key metrics used to assess the tracking stability of each point:\vspace{-0.8em}
\begin{itemize}
    % \item \textbf{Visibility (\%)} – Percentage of frames in which the point was successfully detected. Higher values imply more consistent visibility.
    \item Max gap length – The longest sequence of consecutive frames during which the point was not detected. Smaller gaps indicate better continuity. \vspace{-0.5em}
    % \begin{equation}
    %     G_{max,i} = \max_{g \in G_i} \
    %      \label{eq:max_gap_length}
    % \end{equation}
    % where $G_i$ is the set of all visibility gaps for point $i$, and a gap $g$ is defined as the number of consecutive frames where the point is not visible.
    \item Position variance – The statistical variance in the keypoint’s position, providing a raw measure of how much its location changes over time.\vspace{-0.5em}
    % \begin{equation}
    %     \sigma_i^2 = \frac{1}{N_i}\sum_{t=1}^{T} (p_{i,t} - \bar{p_i})^2
    %     \label{eq:position_variance}
    % \end{equation}
    % where $p_{i,t}$ is the position at frame $t$, $\bar{p_i}$ is the mean position, and $N_i$ is the number of visible frames.
    \item Drift score – A normalized indicator of erratic movement; lower values signify that the keypoint remains stable in space.\vspace{-0.5em}
    % \begin{equation}
    %     D_i = \frac{\sigma_i^2}{\max_{j}(\sigma_j^2)}
    %     \label{eq:drift_score}
    % \end{equation}
    % where $\sigma_i^2$ is the variance of point $i$'s position across visible frames.
    \item Stability category – A high-level classification based on the combination of all metrics, categorizing each point as \textit{Stable}, \textit{Moderately Stable}, \textit{Drifting}, \textit{Frequently Occluded}, or \textit{Occluded}.
\end{itemize}
\begin{table}
    \centering
    \caption{Keypoints Tracking Stability Metrics Across Free-Fall Sequences (Total: 140 frames)}
    % \renewcommand{\arraystretch}{1.2}
    % \scriptsize
    \begin{tabular}{cccccc}
        \hline
        \multirow{2}{*}{\textbf{Keypoints}} & \multirow{2}{*}{\textbf{Point name}}  & \textbf{Max gap} & \textbf{{Drift}} & \textbf{{Position}} & {\textbf{Stability}} \\
         & & \textbf{length (frame)} & \textbf{score (0/1)} & \textbf{variance} & \textbf{category} \\
        \hline
        1 & Neck & 60 & 1.00 & 9.50 & occluded \\
        2 & Eye\_Left &  59 & 0.50 & 0.16 & occluded \\
        3 & Eye\_Right & 54 & 0.51 & 1.27 & occluded \\
        4 & Mouth\_Front\_Top & 56 & 0.06 & 0.14 & occluded \\
        5 & Mouth\_Front\_Bottom &  60 & 0.33 & 5.63 & occluded \\
        6 & Mouth\_Back\_Right & 56 & 0.52 & 10.69 & occluded \\
        7 & Mouth\_Back\_Left &  60 & 0.11 & 2.07 & occluded \\
        8 & Wrist\_Right &  60 & 0.47 & 9.62 & occluded \\
        9 & Wrist\_Left & 51 & 1.00 & 27.35 & occluded \\
        10 & Elbow\_Right &  46 & 0.79 & 26.68 & occluded \\
        11 & Elbow\_Left &  60 & 0.24 & 11.74 & occluded \\
        12 & Shoulder\_Right & 47 & 0.63 & 22.95 & occluded \\
        13 & Shoulder\_Left &  60 & 0.24 & 11.88 & occluded \\
        14 & Torso\_Mid\_Back &  60 & 0.16 & 7.74 & occluded \\
        15 & Ankle\_Right &  60 & 0.50 & 65.20 & occluded \\
        16 & Ankle\_Left & 45 & 0.68 & 45.92 & occluded \\
        17 & Knee\_Right &  46 & 0.63 & 40.96 & occluded \\
        18 & Knee\_Left &  52 & 0.54 & 121.34 & occluded \\
        19 & Hip\_Right &  52 & 0.12 & 19.44 & occluded \\
        20 & Hip\_Left &  52 & 0.55 & 639.61 & occluded \\
        21 & Tail\_Top\_Back &  60 & 0.01 & 16.92 & occluded \\
        22 & Tail\_Mid\_Back &  60 & 0.21 & 9.86 & occluded \\
        23 & Tail\_End\_Back &  60 & 0.15 & 189.43 & occluded \\
        \hline
    \end{tabular}
    \label{tab:keypoint_stability}
\end{table}\vspace{-1em}
\begin{figure}
    \centering
    \includegraphics[width=0.9\linewidth]{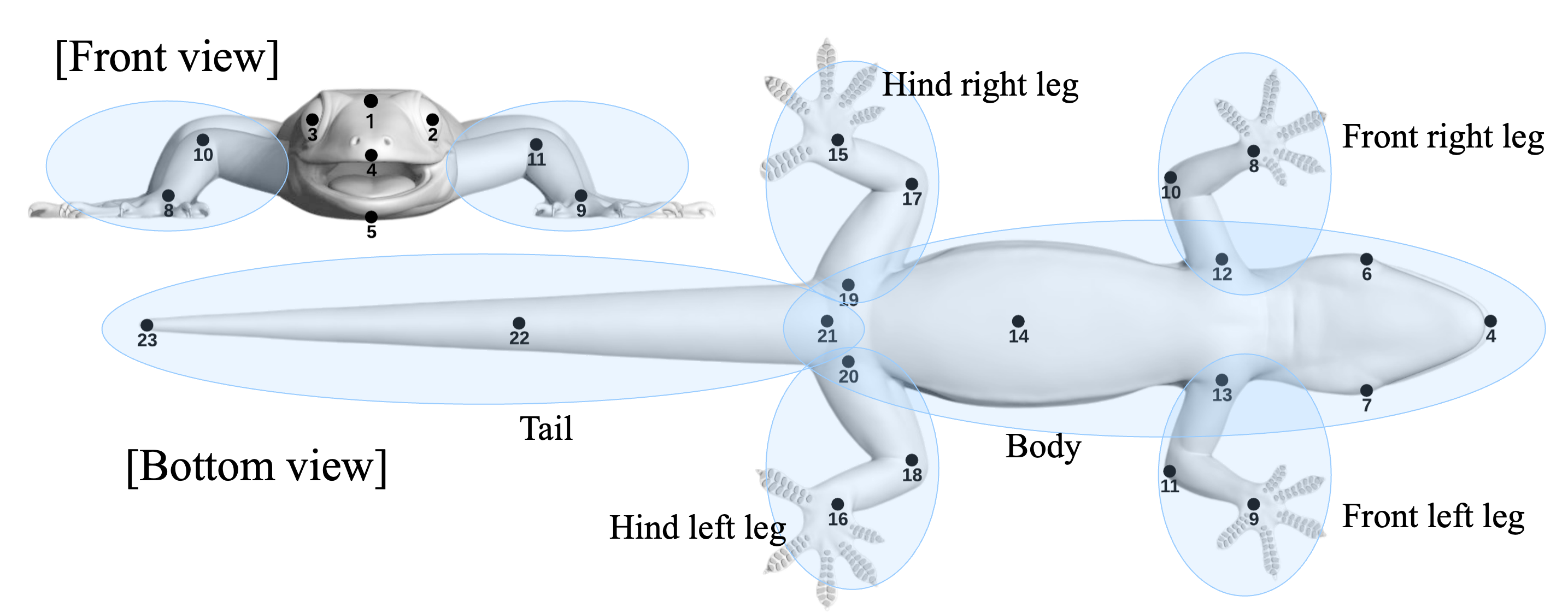}
    \caption{Selected 6 Primary Anatomical Body Segments}
    \label{fig:lizard_segment}
\end{figure}

The stability analysis revealed that most keypoints suffered from frequent occlusion or drift during the observed free-fall sequences. As shown in Table~\ref{tab:keypoint_stability}, all points were ultimately categorized as \emph{Occluded}, highlighting the challenges of reliable tracking under dynamic conditions. Despite this, certain keypoints, such as 19, 21, and 23, exhibited comparatively lower drift scores and moderate visibility, suggesting partial robustness.

Of particular interest for control strategies in SMSs are points with low drift scores (e.g., keypoint 21 at 0.01 and keypoint 19 at 0.12), which may serve as inertial references or anchors in future reconstruction frameworks. However, the prevalence of long occlusion gaps (up to 60 consecutive frames) and limited visibility across the board emphasizes the need for either improved multi-camera coverage or refined prediction models to enhance reliability. 

Based on this comprehensive evaluation, which includes movement magnitude, visibility, temporal continuity, and drift behavior, a reduced set of 11–17 
keypoints is selected, excluding those around the eyes and mouth. These keypoints are grouped into \emph{six primary segments}: the \emph{body}, \emph{tail}, and \emph{four legs}, as highlighted in blue in Figure~\ref{fig:lizard_segment}. 

This selection prioritizes keypoints that are both biomechanically relevant and statistically stable across frames, ensuring robustness against occlusion and erratic motion. Note that each body segment, such as the body, tail, and four legs, can be represented as the robotic arm, base spacecraft, and actuators, respectively. Although the legs display relatively small movements during self-righting, their influence on the body’s motion trajectory warrants further analysis in conjunction with Step 3 to refine the biomechanical model.
The proposed configuration balances model simplicity with tracking reliability, reduces noise in control signals, and aligns with the structural and dynamic requirements of SMS operating in microgravity environments.
% \begin{figure}[tb]
%     \centering
%     \includegraphics[width=0.9\linewidth]{Figures/fig_lizard_segment_rev.png}
%     \caption{Selected 6 Primary Anatomical Body Segments}
%     \label{fig:lizard_segment}
% \end{figure}

\subsubsection{\textbf{Body Segment Motion Extraction and Analysis}}
To analyze the biomechanical contributions of each anatomical segment during aerial righting, this study conducts a quantitative analysis of the angular motions of the body, tail, and limbs by transforming 2D motion data from image frames into 3D motion in the real world. That is, one evaluates segmental motions, such as (i) the orientation of each segment with respect to the inertial frame and (ii) the relative orientation of the limbs relative to the body. 

\paragraph{Motion of Each Body Segment with respect to the Inertial Frame}
In the first stage, motions of all segments, including the body, tail, and limbs, are extracted with respect to the inertial frame. The 3D positions of anatomical landmarks (or keypoints), reconstructed from synchronized multi-camera video, are used to define rigid body segments as shown in Figure~\ref{fig:lizard_segment}. Each segment is assigned a local coordinate frame, fixed on the respective body segment, constructed from anatomically meaningful keypoints. 

For the main body, the $x$-axis unit vector ($\hat{\bf x}_b$) of the body-fixed reference frame is primarily defined as the normalized vector from the vent (keypoint 21) to the neck (keypoint 1). The temporary $y$-axis unit vector ($\hat{\bf y}_{b_\text{temp}}$), which generates the $x$-$y$ plane, is formed using the vector from the right shoulder (keypoint 12) to the left shoulder (keypoint 13). After defining the $z$-axis unit vector ($\hat{\bf z}_b$) as the vector perpendicular to the {$x$-$y$} plane ($\hat{\bf z}_b=\hat{\bf x}_b\times \hat{\bf y}_{b_\text{temp}}$), the $y$-axis unit vector ($\hat{\bf y}_b$) is then determined using the right-hand rule. Based on these unit vectors, the direction cosine matrix (DCM) representing the rotation of the body reference frame with respect to the inertial frame is defined as:
\begin{equation}\label{eq:dcm}
    C_{BN} = \begin{bmatrix} \hat{\bf x}_b^T \\ \hat{\bf y}_b^T \\ \hat{\bf z}_b^T  \end{bmatrix}.    
\end{equation}

Similarly, the tail reference frame is defined using the tail tip (keypoint 23) and vent (keypoint 21) for the $x$-axis unit vector ($\hat{\bf x}_t$) and the hip joints (keypoints from 20 to 19) for the temporary $y$-axis unit vector ($\hat{\bf y}_{t_\text{temp}}$). Thus, the DCM for the tail motion ($C_{TN}$) is obtained following the same procedure as for the body-fixed reference frame.

\begin{figure}[tbp]
    \centering
    \includegraphics[width=0.8\linewidth]{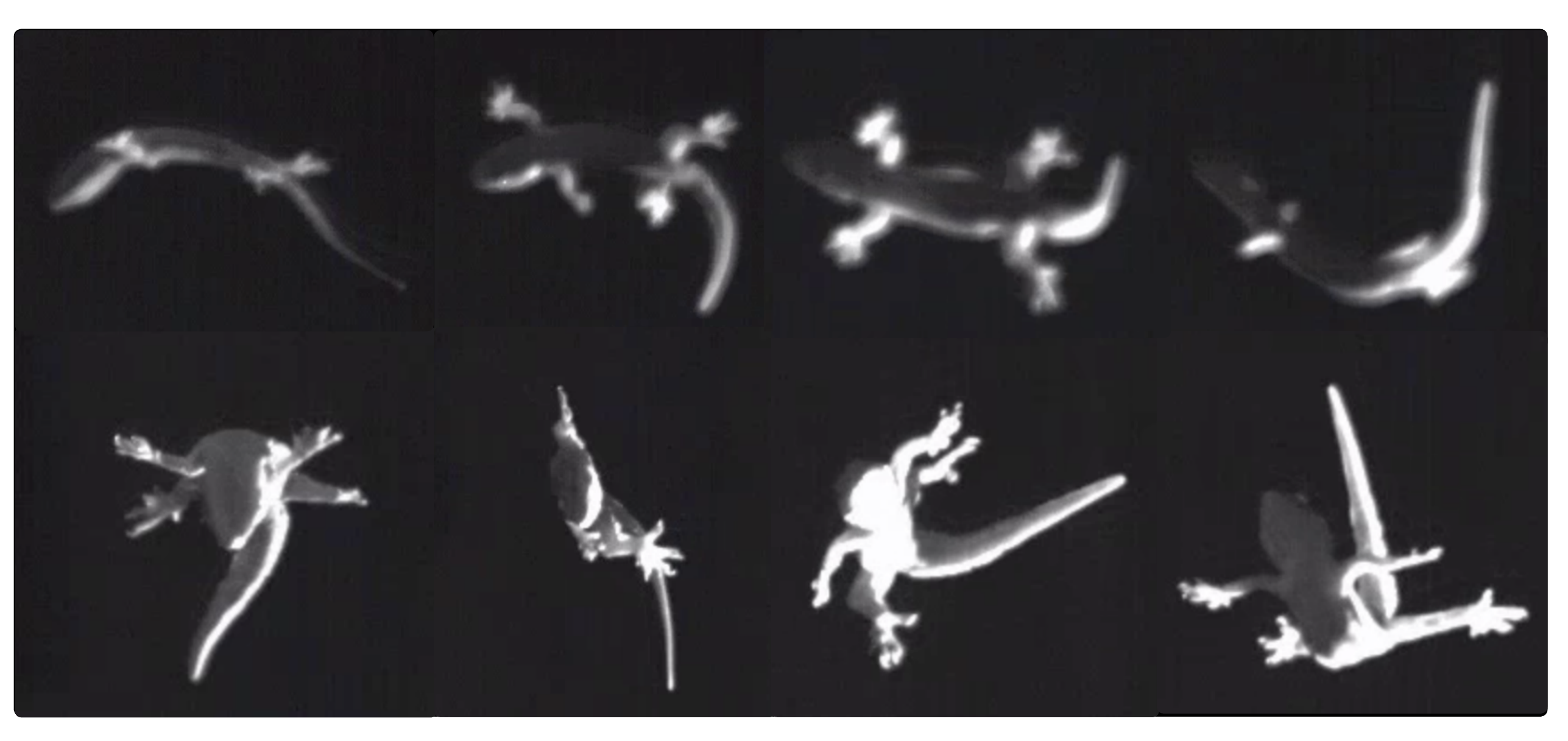}
    \caption{Sequences of Righting Reflex Motions (Top: side view, Bottom: front view)}
    \label{fig:lizard_sequence}%\vspace{-1.5em}
\end{figure}

For limb segments, local reference frames are primarily constructed using the leg-to-body vector as the $y$-axis: from the wrist (keypoints 8 (front) and 15 (hind)) to the shoulder (keypoint 12 (front)) and hip (keypoint 19 (hind)) for right legs, and from the shoulder (keypoint 13 (front)) and hip (keypoint 20 (hind)) to the wrist (keypoint 9 (front) and 16 (hind)) for the left legs. Note that the $y$-axis unit vector is expressed as $\hat{\bf y}_{l_i}$, where the subscript $l_i$ represents left front, left hind, right front, and right hind legs, respectively. 

Next, the $z$-axis unit vector ($\hat{\bf z}_{l_i}$) for each leg is defined as a vector that is orthogonal to the {$x$-$y$} plane produced by $\hat{\bf y}_{l_i}$ and the {$x$}-axis of the inertial frame. Then, the {$x$}-axis unit vector is determined as $\hat{\bf x}_{l_i} = \hat{\bf y}_{l_i} \times \hat{\bf z}_{l_i}$. Finally, the DCM for the motion of each leg with respect to the inertial frame ($C_{L_i N}$) is derived, similar to Eq.~\eqref{eq:dcm}.

\begin{figure}
    \centering
    \begin{subfigure}[b]{0.49\textwidth}
        \includegraphics[width=\linewidth]{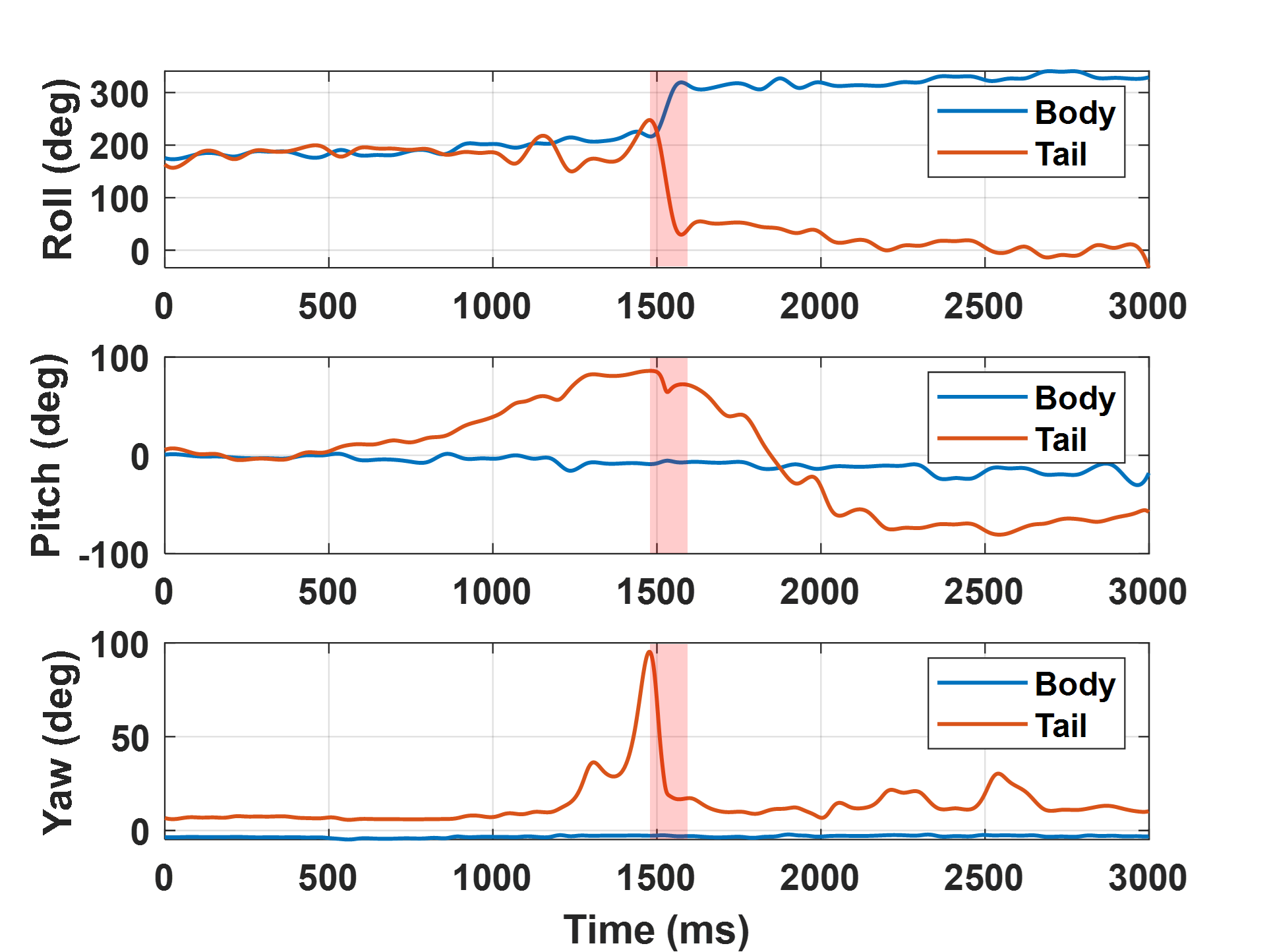}
        \caption{Body and Tail (Entire Video)}
        \label{fig:body_tail_entire}
    \end{subfigure}
    \hfill
    \begin{subfigure}[b]{0.49\textwidth}
        \includegraphics[width=\linewidth]{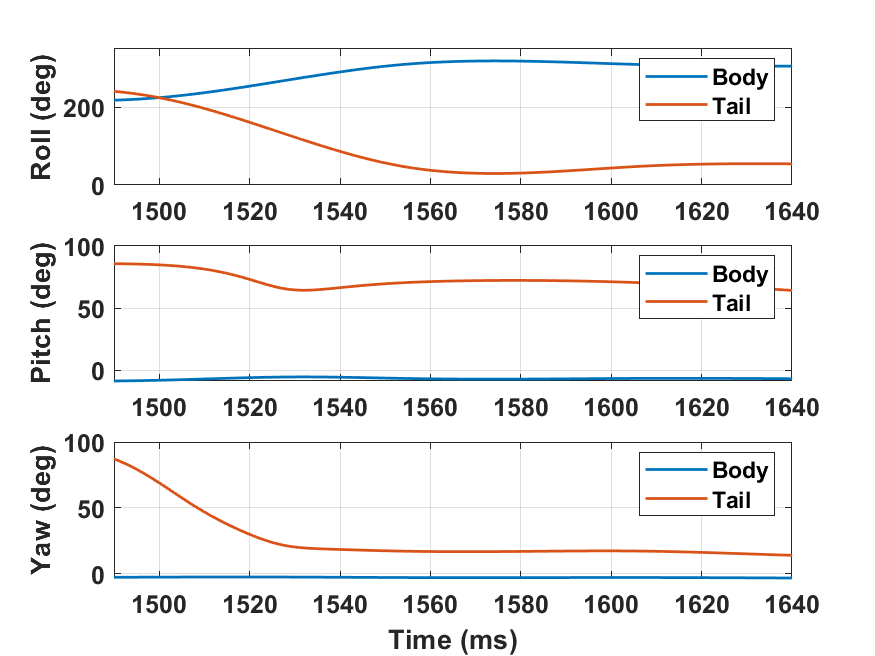}
        \caption{Body and Tail (During Righting)}
        \label{fig:body_tail_righting}
    \end{subfigure}

    \begin{subfigure}[b]{0.49\textwidth}
        \includegraphics[width=\linewidth]{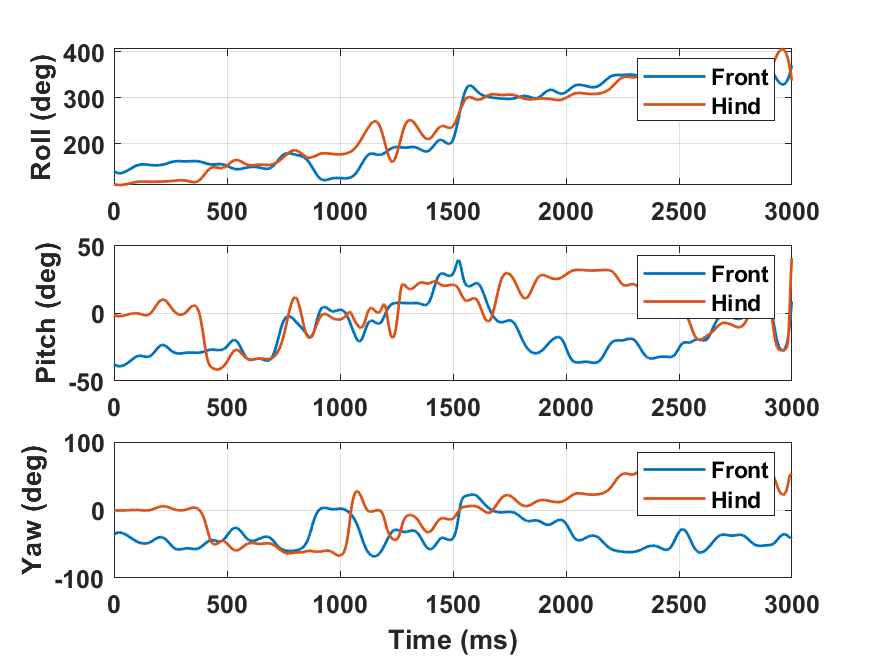}
        \caption{Left Legs (Entire Video)}
        \label{fig:left_entire}
    \end{subfigure}
    \hfill
    \begin{subfigure}[b]{0.49\textwidth}
        \includegraphics[width=\linewidth]{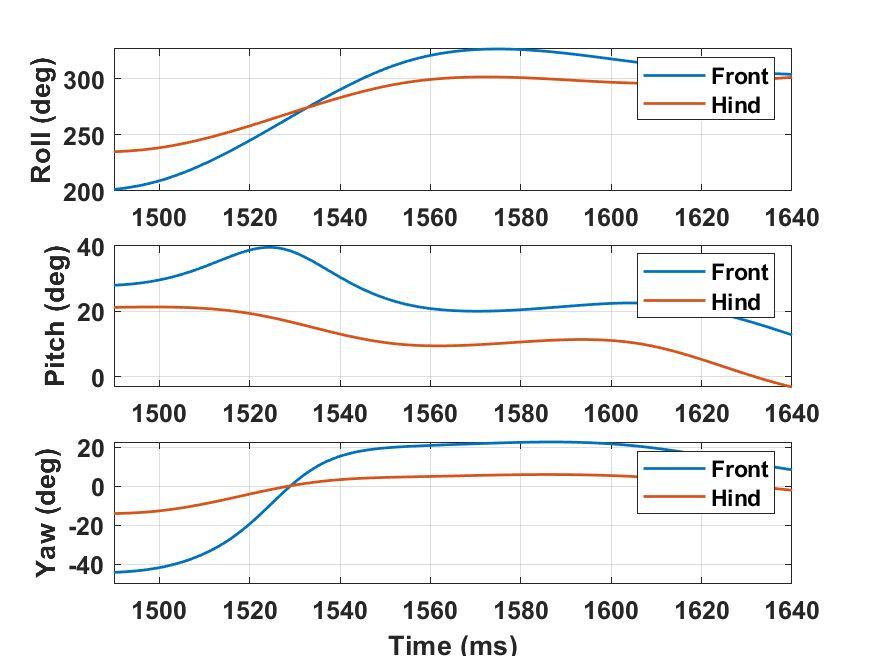}
        \caption{Left Legs (During Righting)}
        \label{fig:left_righting}
    \end{subfigure}

    \begin{subfigure}[b]{0.49\textwidth}
        \includegraphics[width=\linewidth]{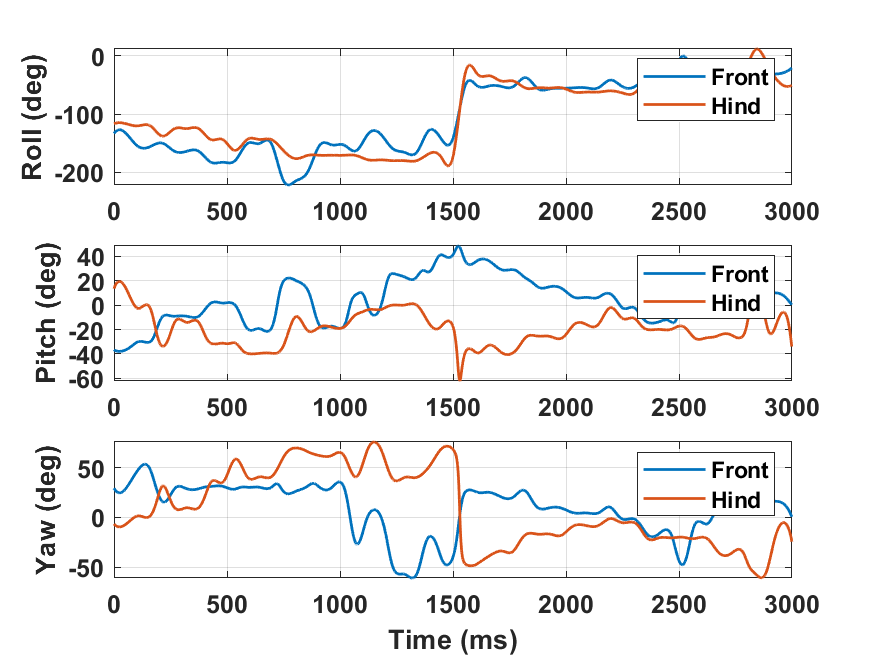}
        \caption{Right Legs (Entire Video)}
        \label{fig:right_entire}
    \end{subfigure}
    \hfill
    \begin{subfigure}[b]{0.49\textwidth}
        \includegraphics[width=\linewidth]{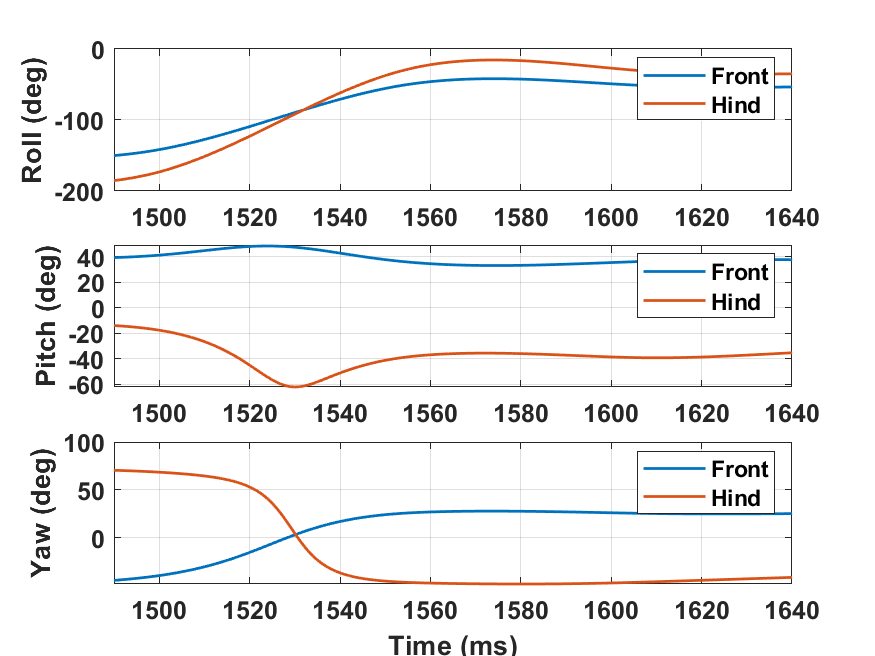}
        \caption{Right Legs (During Righting)}
        \label{fig:right_righting}
    \end{subfigure}
    
    \caption{Body Segment Motion with respect to the Inertial Frame}
    \label{fig:lizard_motion}%\vspace{-1.5em}
\end{figure}

While Figure~\ref{fig:lizard_sequence} illustrates the primary sequences of the lizard's righting reflex motions obtained from the free-fall video, Figure~\ref{fig:lizard_motion} shows the time-aligned attitude trajectories of the lizard's body segments, which include the entire motion trajectory of the experimental video. To concentrate on the righting reflex of the lizard, the motion data for the specific time duration between 1,490 and 1,640 ms, highlighted in red in Figure~\ref{fig:body_tail_entire}, is displayed on the right side of Figure~\ref{fig:lizard_motion}. Note that the attitude information is represented by the 3-2-1 set of the Euler angles, which are converted from the DCMs of each body segment. 

Since the direction from the vent to the neck aligns with the body's roll direction, the primary motion changes in the body and tail during righting are observed along the roll axis, indicating aerial righting due to the transferred momentum from the tail rotation. 
Additionally, the extracted angular motions of each leg, particularly along the roll axis, shown in Figures~\ref{fig:left_entire}--~\ref{fig:right_righting}, follow a similar trend to the body's roll motion, with increasing angles. 

However, the roll angle at 1,640 ms in Figure\ref{fig:body_tail_righting} is less than 360 deg, which %seems not to be a complete body rotation, while it seems that 
suggests that the body has not completed a full rotation, despite the righting sequence in Figure~\ref{fig:lizard_segment} showing a full rotation. While the primary righting reflex is achieved between 1,490 and 1,640 ms, additional body posture adjustments are observed after righting, as shown in Figure~\ref{fig:body_tail_entire}. 
Furthermore, the video's low-quality images in each frame may cause such a discrepancy, reducing accuracy. %For this reason
To address this, our research group is currently building a test environment for the lizards' righting reflex using multiple high-speed cameras with different views, ensuring high-quality video capture for improved motion extraction accuracy.

% The corresponding motions are displayed as screenshots shown in Figure~\ref{fig:lizard_sequence}.
% Additionally, Figure~\ref{fig:leg_righting} depicts the extracted angular motion of each leg. Similar to the body's roll trajectory, the legs' roll orientations also have increasing trends. 

% \begin{figure}[t]
%     \centering
%     \begin{subfigure}[b]{0.49\textwidth}
%         \includegraphics[width=\linewidth]{Figures/fig_body_tail_entire_.png}
%         \caption{Entire Video}
%         \label{fig:body_tail_entire}
%     \end{subfigure}
%     \hfill
%     \begin{subfigure}[b]{0.49\textwidth}
%         \includegraphics[width=\linewidth]{Figures/fig_body_tail_righting.png}
%         \caption{During Righting}
%         \label{fig:body_tail_righting}
%     \end{subfigure}
%     \caption{Body and Tail Motion with respect to the Inertial Frame}
%     \label{fig:body_tail}%\vspace{-1.5em}
% \end{figure}
% \begin{figure}
%     \centering
%     \begin{subfigure}[b]{0.49\textwidth}
%         \includegraphics[width=\linewidth]{Figures/fig_left_legs_righting.png}
%         \caption{Left Legs}
%         \label{fig:left_righting}
%     \end{subfigure}
%     \hfill
%     \begin{subfigure}[b]{0.49\textwidth}
%         \includegraphics[width=\linewidth]{Figures/fig_right_legs_righting.png}
%         \caption{Right Legs}
%         \label{fig:right_righting}
%     \end{subfigure}
%     \caption{Legs' Motion with respect to the Inertial Frame During Righting}
%     \label{fig:leg_righting}%\vspace{-0.5em}
% \end{figure}

% \begin{figure}
%     \centering
%     \begin{subfigure}[b]{0.49\textwidth}
%         \includegraphics[width=\linewidth]{Figures/fig_left_legs_righting.png}
%         \caption{Left Legs}
%         \label{fig:left_righting}
%     \end{subfigure}
%     \hfill
%     \begin{subfigure}[b]{0.49\textwidth}
%         \includegraphics[width=\linewidth]{Figures/fig_right_legs_righting.png}
%         \caption{Right Legs}
%         \label{fig:right_righting}
%     \end{subfigure}
%     \caption{Legs' Motion with respect to the Inertial Frame During Righting}
%     \label{fig:leg_righting}%\vspace{-0.5em}
% \end{figure}

\paragraph{Relative Leg Motion with respect to the Body}

As illustrated, the motion changes in each leg are also observed. Since the legs are connected to the body, it is required to explore the relative motion of the legs with respect to the body to analyze the impact of leg motion on the righting reflex. 

The relative rotation of each leg with respect to the body, defined as $C_{L_i B}$ ($L_i$ represents left front, left hind, right front, and right hind legs), is obtained by the successive rotations from the body to each leg, defined as 
\begin{equation}
    C_{L_i B} = C_{L_i N}C_{BN}^T.
\end{equation}

\begin{figure}
    \centering
    \begin{subfigure}[b]{0.49\textwidth}
        \includegraphics[width=\linewidth]{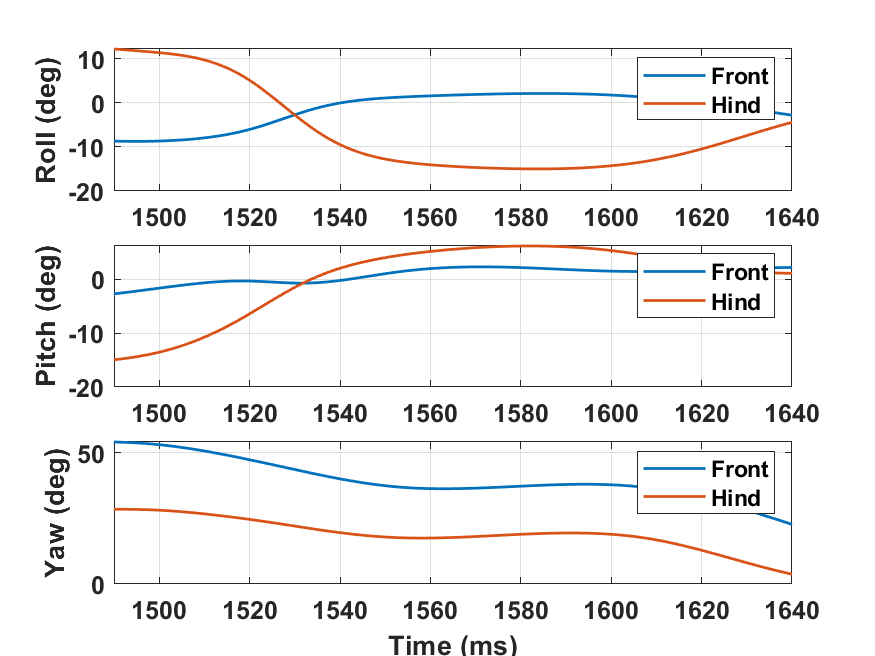}
        \caption{Left Legs}
        \label{fig:left_relative}
    \end{subfigure}
    \hfill
    \begin{subfigure}[b]{0.49\textwidth}
        \includegraphics[width=\linewidth]{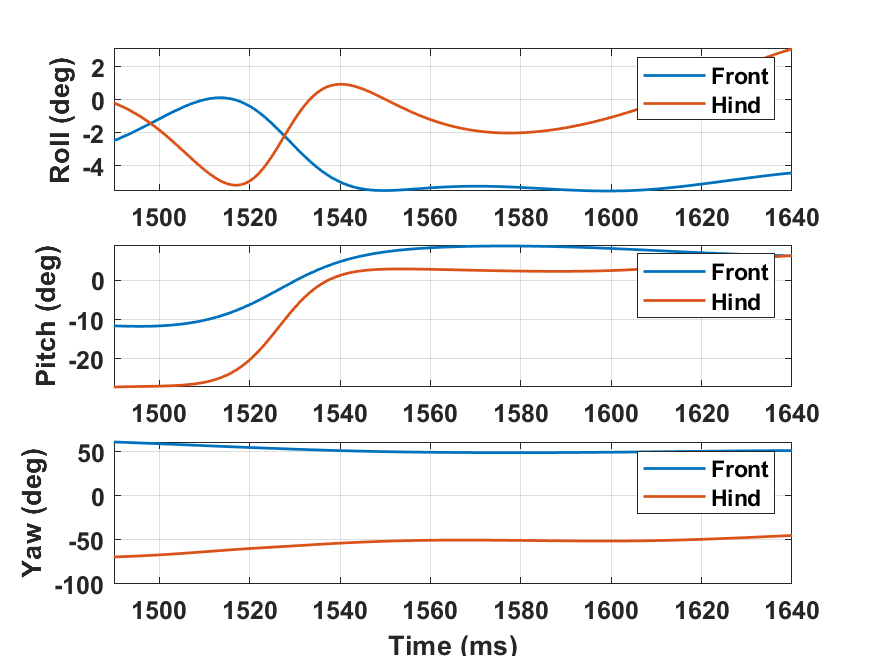}
        \caption{Right Legs}
        \label{fig:right_relative}
    \end{subfigure}
    \caption{Legs' Relative Motion with respect to the Body During Righting}
    \label{fig:leg_relative}
\end{figure}

Figure~\ref{fig:leg_relative} illustrates the relative angular trajectories of each leg relative to the body. One can observe that the relative rolling motion changes, which is the primary motion during the righting reflex, remain small, around the zero angle, approximately within $\pm$ 15 deg. These findings suggest that the tail provides the dominant control for body posture stabilization during mid-air righting maneuvers, while the legs function as passive stabilizers, contributing negligibly to body motion. 

This insight could inform the design of an energy-efficient SMS operational strategy, where leg actuation is reflected in the operation of the base spacecraft's attitude control actuators, which may help aid and/or correct the minor base motion changes, improving overall performance.

\subsection{Integration and Control Evaluation in SMS (In progress)}
The tasks in this step include analyzing the objectives and weighting factors for the optimization framework (defined in the first step) using the extracted motion trajectories. Then, these trajectories are leveraged as reference trajectories for the SMS, 
allowing the influence of different weights to be analyzed by observing the SMS's behaviors. However, due to the limited lizard's motion data available, the current work focuses primarily on implementing the extracted motion trajectory into the SMS.

\subsubsection{\textbf{Lizard's Body and Tail Motion Trajectory Analysis During Righting}}

Given the similarities between the lizards and SMSs, the lizard's tail motion is primarily utilized as the reference input trajectory for the SMS. To confirm that the lizard's motion is applicable to SMSs, the orientation and rate along the roll axis, which is the primary axis for the righting reflex, are computed. 

Note that the rate information is computed based on numerical differentiation using the orientation data. The entire maneuver occurs within a span of 150 ms, which results in a very high rotation rate of approximately 2,000 -- 4,000 deg/s. Since %the 
such rapid movement is impractical for SMSs, the given trajectories (spanning 150 ms) are scaled to those with 225 s, which is more suitable for implementation on SMSs. 
It is important to note that the scaled time is based on the operational speed of the ETS-VII's robotic arm, which operates at approximately within 5 deg/s \cite{Abiko2001}.

\begin{figure}[tbp]
    \centering
    \includegraphics[width=0.5\linewidth]{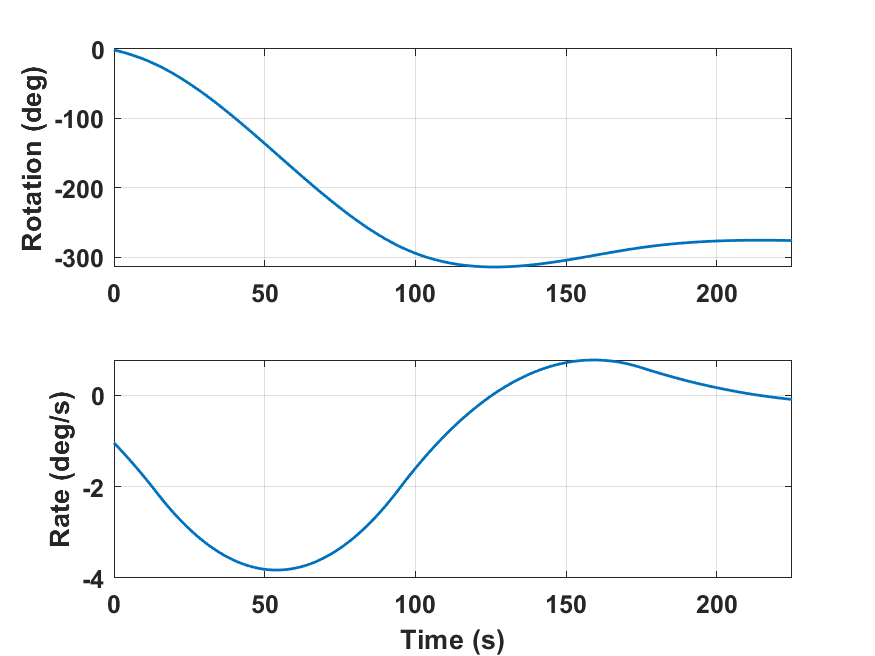}
    \caption{Scaled Relative Tail Motion with respect to the Body}
    \label{fig:scaled_motion}
\end{figure}

Using the scaled trajectory, the relative tail rotation with respect to the body is selected as the reference joint trajectory for the robotic arm of the SMS, as displayed in Figure~\ref{fig:scaled_motion}. This is because the joint motion of the SMS is typically described with respect to the spacecraft base frame.
The characteristic information of this step response, assuming it reaches steady-state at 225 s, is summarized as follows: 
a rise time of 64.5 s, settling time of 169.5 s (within 5 \% of the final value), and overshoot of 13.85 \%. This motion trajectory, with these characteristics, will be implemented into the SMS as a reference trajectory for the first joint of the robotic arm.
% \begin{figure}[h]
%     \centering
%     \begin{subfigure}[b]{0.48\textwidth}
%         \includegraphics[width=\linewidth]{Figures/fig_lizard_roll_motion.png}
%         \caption{Original Trajectory}
%         \label{fig:lizard_original}
%     \end{subfigure}
%     \hfill
%     \begin{subfigure}[b]{0.48\textwidth}
%         \includegraphics[width=\linewidth]{Figures/fig_scaled_lizard_motion.png}
%         \caption{Scaled Trajectory}
%         \label{fig:lizard_scaled}
%     \end{subfigure}
%     \caption{Lizard's Body and Tail Rotation and Rate along with the Roll Axis}
%     \label{fig:lizard_trajectory}
% \end{figure}
% \begin{figure}[h]
%     \centering
%     \includegraphics[width=0.5\linewidth]{Figures/fig_relative_tail_motion.png}
%     \caption{Scaled Relative Tail Motion with respect to the Body}
%     \label{fig:relative_tail}
% \end{figure}
%In this motion response, assuming that it reaches the steady state value at 225 s, the 

\subsubsection{\textbf{Trajectory Implementation to SMS}}
\begin{figure}[t]
    \centering
    \includegraphics[width=0.4\linewidth]{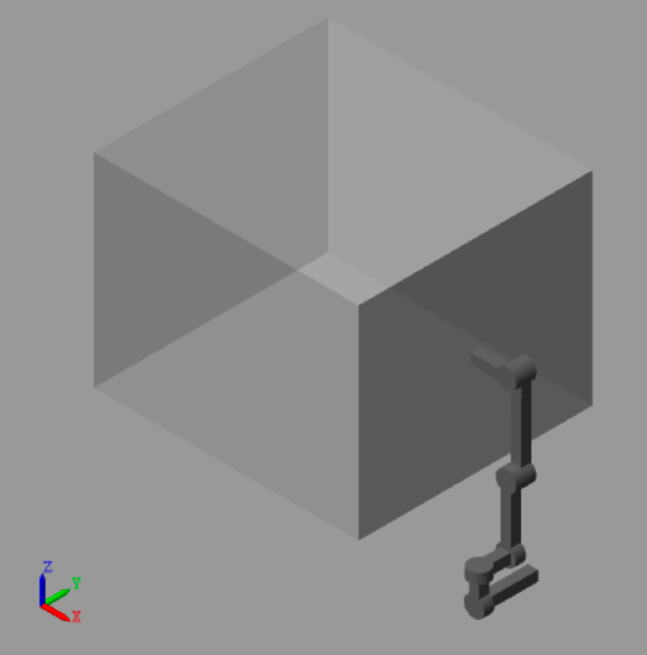}
    \caption{Simplified SMS Model Based on ETS-VII Specifications}
    \label{fig:simplified_model}
\end{figure}

To implement the lizard's motion trajectory into SMSs, this work considers an actual SMS model, ETS-VII, which has been tested in space. Based on the available specifications \cite{yoshida2003ets}, a simplified SMS model is built as illustrated in Figure~\ref{fig:simplified_model}. It is worth noting that the mass, inertia, and dimensions are identical to ETS-VII's specifications, while the location of the robotic arm's first link has been modified to align the first link's rotation axis with the roll axis, similar to the alignment of the lizard frames in the motion extraction step. Moreover, the initial roll orientation of the base and first joint is set to around 180 deg to observe a motion trajectory similar to that of the lizard.

\begin{figure}[h]
    \centering
    \includegraphics[width=0.5\linewidth]{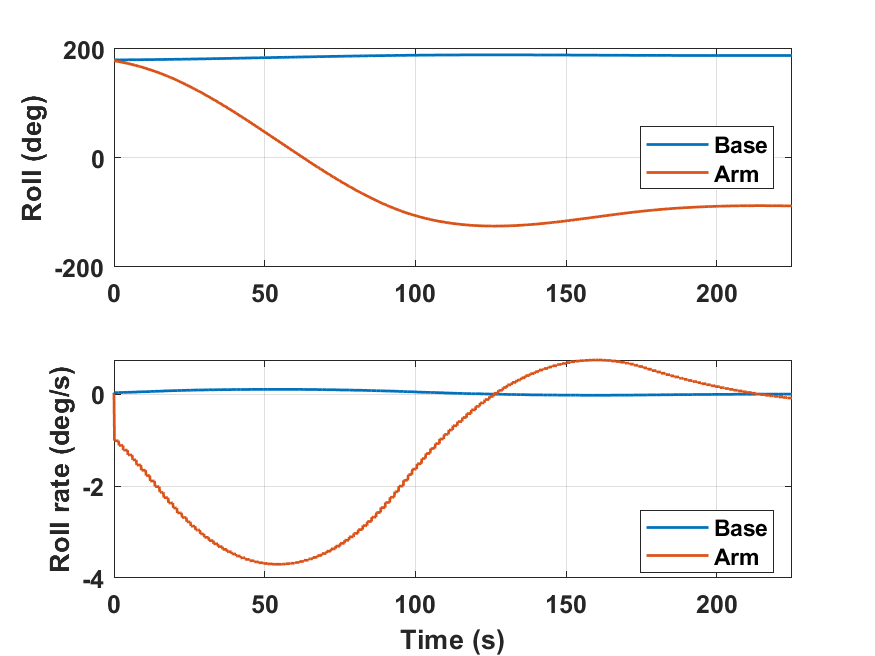}
    \caption{Lizard's Scaled Trajectory-Implemented SMS Motion}
    \label{fig:resultant_sms_motion}
\end{figure}

With this setting, the relative tail motion trajectory in Figure~\ref{fig:scaled_motion} is applied to the first joint, and the resultant SMS behavior is illustrated in Figure~\ref{fig:resultant_sms_motion}. Although the arm's joint trajectory trend mirrors the relative tail trajectory, small motion changes in the base (orientation: less than 10 deg, rate: less than 0.15 deg/s) are observed. This is expected because the base inertia is much greater than that of the robotic arm. In fact, if the base's roll axis inertia is reduced by approximately 1/20, the SMS's roll axis motion can be made to closely resemble the lizard's roll motion trajectory shown in Figure~\ref{fig:body_tail_righting}.
% For this reason, the base's roll axis inertia is intentionally reduced by 1/20 to obtain the similar behavior from the SMS, and the resultant behavior with the modified body inertia is shown in Figure~\ref{fig:sms_motion_modified_inertia}. 
% As we can expect, similar to the lizard's righting motion, the base's roll orientation and its rate are opposite to the arm's motion, and the overall trajectories are similar to the ones of the lizard (Figure~\ref{fig:lizard_scaled}). 

The different motion behaviors when implementing the lizard trajectory into the SMS are primarily due to the differences in inertia ratios between the lizard and SMS, as listed in Table~\ref{tab:ratio_comp}, although they share a similar mass ratio. 
The inertia associated with the tail (or arm), which drives the body (or base) movement, is influenced by mass distribution, especially for the center of mass (CM) location. 
While the length ratio is not explicitly included in this comparison, it is implicitly captured in the CM calculation.

As indicated in the table, reducing the SMS's roll axis inertia by 1/20 brings its inertia ratio closer to that of the lizard. This suggests that the mass and inertia ratio, considering mass distribution, are the dominant factors that contribute to the body rotation during righting.

% \begin{figure}
%     \centering
%     \begin{subfigure}[b]{0.48\textwidth}
%         \includegraphics[width=\linewidth]{Figures/fig_sms_motion2.png}
%     \caption{with Original ETS-VII Properties}
%     \label{fig:resultant_sms_motion}
%     \end{subfigure}
%     \hfill
%     \begin{subfigure}[b]{0.48\textwidth}
%         \includegraphics[width=\linewidth]{Figures/fig_sms_scaled_motion.png}
%     \caption{with the Modified Inertia}
%     \label{fig:sms_motion_modified_inertia}
%     \end{subfigure}
%     \caption{Lizard's Scaled Trajectory-Implemented SMS Base and Arm Motion}
%     \label{fig:sms_motion}
% \end{figure}

As shown, direct implementation of the lizard's motion into the SMS is possible, but impractical due to discrepancies, such as the inertia ratio, as well as the differing objectives behind the motion changes. However, it is still too early stage to conclude this analysis, as this work considers only one case of the lizard's motion due to a lack of experimental data. Therefore, further work is required to analyze not only the lizard's objectives and the combinations of the weighting factors but also to understand their behavior trends along with the step response characteristics. Additionally, analyzing relationships or correlations among the response characteristics, the mass and inertia ratios, and the relative importance of the objectives introduced in Eq. \eqref{eq:cost} is needed for a more in-depth analysis.

To remedy this, our research group is currently setting up an experimental environment for the lizard's righting reflex tests, which will allow for the acquisition of more lizard motion data during righting using multiple high-speed cameras. Once these experimental data are obtained and analyzed, this step will be validated.

\begin{table}
    \centering
    \caption{Mass and Inertia Ratio Comparison}
    \label{tab:ratio_comp}
    \begin{tabular}
    {>{\raggedright\arraybackslash}p{7.5cm}
    >{\centering\arraybackslash}p{2.5cm}
    >{\centering\arraybackslash}p{3cm}}
    \hline
        \textbf{{Parameter}} & \textbf{Lizard}\cite{jusufi2008active} & \textbf{SMS (ETS-VII)}\cite{yoshida2003ets} \\ \hline
        Mass ratio (tail / body or arm / base) & 0.1 & 0.055 \\ 
        $\ \ $ Body or base mass (kg) & 2.9$\times$10$^{-3}$ kg & 2550 kg \\ 
        $\ \ $ Tail or arm mass (kg) & 0.29$\times$10$^{-3}$ kg & 140.4 kg \\ \hline
        Inertia ratio (tail / body or arm / base)  & 0.8 & 0.056* \\ 
        $\ \ $ Body or base inertia (kg-m$^2$) & 6.6$\times$10$^{-8}$ & 6200 (roll axis) \\ 
        $\ \ $ Tail or arm inertia governed by CM** (kg-m$^2$) & 5.29$\times$10$^{-8}$ & 360  \\ \hline
        \multicolumn{3}{l}{\footnotesize * When the base inertia is adjusted by 1/20, the inertia ratio is changed to 0.86, resulting in similar behaviors.} \\ 
        \multicolumn{3}{l}{\footnotesize ** The effective inertia affecting the body or base rotation is the one governed by the CM (center of mass) location.}
    \end{tabular}
\end{table}

\section{Conclusions}

This work presents a biologically inspired approach for space manipulator system (SMS) operations by translating lizard righting reflex behaviors into reference motion trajectories for SMSs. Through a structured analysis, this study establishes morphological, behavioral, and environmental analogies demonstrating the feasibility of leveraging evolved motion patterns of lizards for SMS operations. Moreover, using a computer vision pipeline, air-righting motion trajectories are extracted and analyzed, confirming that tail dynamics primarily drive reorientation for body posture stabilization while limb motions contribute to secondary stabilization or minor motion change. The extracted lizard's motion trajectory is then implemented into the SMS, observing its feasibility and identifying discrepancies between the lizards and SMSs. These insights support the formulation of interpretable multi-objective optimization problem settings balancing stability, safety, and efficiency. 
Limitations include the small dataset and the need to appropriately rescale motion profiles for practical SMS implementation. Future work will acquire lizard experimental motion data to validate the final step.
Overall, this study provides the basis for incorporating biological principles into robust and interpretable control architectures for space robotics.

\section{Acknowledgement}
This material is based upon work supported by the Air Force Office of Scientific Research under award number FA9550-24-1-0600.

\bibliographystyle{AAS_publication}   
\bibliography{references}   

\end{document}